%% file: acl2023.tex
\pdfoutput=1

\documentclass[11pt]{article}

\usepackage[]{ACL2023}

\usepackage{times}
\usepackage{latexsym}
\usepackage{graphicx}
\usepackage{booktabs}
\usepackage{multirow}
\usepackage{lipsum}   
\usepackage{tcolorbox}
\usepackage[table]{xcolor}

\usepackage[T1]{fontenc}

\usepackage[utf8]{inputenc}

\usepackage{microtype}

\usepackage{inconsolata}

\title{The Shape of Power: A Multilingual Framework for Social Power Reasoning in Dialogues
}
\author{
  Farah Atif \quad
  Sougata Saha \quad
  Monojit Choudhury \\[2mm]
  Mohamed bin Zayed University of Artificial Intelligence \\
  \texttt{\{farah.atif,sougata.saha,monojit.choudhury\}@mbzuai.ac.ae}
}
\usepackage{CJKutf8}

\begin{document}
\maketitle
\begin{CJK}{UTF8}{mj}
\begin{abstract}

Social power plays a fundamental role in shaping human interaction, yet computational studies of power remain limited to narrow linguistic and cultural settings. Existing datasets further lack the demographic and relational depth needed for robust cross-cultural analysis. To address this gap, we introduce a theoretically grounded framework for studying social power in naturalistic multilingual dialogue through movie screenplays. The framework integrates a schema informed by social science theory, a native speaker annotation pipeline refined through pilot studies, and a custom interface for scalable cross-lingual analysis. Using this framework, we constructed an initial corpus containing 15,836 annotated instances from 100 scenes in French and Egyptian Arabic movies. Our analysis reveals strong agreement on observable demographic and contextual attributes, while socially interpretive aspects, such as power asymmetry and intention alignment, remain more contested, highlighting the complexity of social power across cultures. We evaluated 6 Large Language Models (LLMs) and Multimodal LLMs on cross-cultural social power reasoning, finding persistent gaps between human and model agreement in relational and theory-of-mind reasoning. Our work introduces the first extensible multilingual framework for studying social power in dialogues and provides an initial evaluation setting for studying cross-cultural social reasoning.

\end{abstract}

\section{Introduction}

Power is a foundational construct in social science; it shapes the structure of societies, governs interpersonal relationships, and is deeply embedded in cultural norms and communicative practices \cite{Foucault1980-FOUPSI,Hofstede2011}. From political institutions to everyday conversation, power determines who speaks, who is heard, and whose words carry weight. Because it is simultaneously a social, cultural, and psychological phenomenon, understanding power requires analysis that is both theoretically grounded and empirically broad.

Despite its centrality, power has not been studied systematically across cultures within computational linguistics. Existing work tends to treat power as a monolithic concept tied to a single language or social context, and the conversational corpora available today lack the demographic, relational, and cultural depth needed to study how power is expressed, perceived, and negotiated across societies. As a result, researchers are often constrained to narrow perspectives, specific demographic groups, individual languages, or particular interaction types, which can limit the extent to which findings from one cultural context are validated or generalized to others. 
This leaves several questions fundamentally open: \textit{i) How is power socially manifested and expressed through demographic and interactional features across cultures? ii) Can LLMs reliably identify power dynamics in naturalistic dialog? iii) How does LLM behavior on power-related tasks vary across languages and cultures, and do multimodal models offer additional gains?}
Answering any of these questions is currently intractable, as there is still no principled, scalable way to capture power in naturalistic, cross-cultural interactions, nor sufficiently rich annotated data to support analysis at the required depth.

We address this gap by introducing a framework for the cross-cultural study of social power in naturalistic dialog. The framework consists of three components: i) a theoretically-grounded annotation schema that operationalizes power through demographic, relational, and theory-of-mind variables drawn from established social-science traditions; ii) an annotation pipeline in which trained native speakers produce all annotations, and refined through iterative pilot studies; iii) a custom annotation interface designed around HCI principles for cognitive tractability and cross-lingual extensibility. The framework is language-agnostic by construction, with culturally-specific categories supported where universal taxonomies are insufficient.
To demonstrate the framework, we introduce \textit{SocLens}\footnote{https://github.com/mbzuai-nlp/Social-Power-in-Dialogs}), an initial multilingual dataset spanning two typologically and culturally distinct languages, French and Arabic (Egyptian dialect). \textit{SocLens} contains 100 densely annotated scenes drawn from screenplays grounded in everyday social interactions. Each scene is independently annotated by native speakers, enabling feature-level inter-annotator agreement analysis. We further benchmark state-of-the-art text and multimodal LLMs against the human annotations and analyze their social reasoning behavior.


\input{Chapters/relatedwork.tex}

\input{Chapters/principles}

\section{Dataset Creation}

Figure~\ref{fig:pipeline} presents the overall workflow used for constructing the dataset. The pipeline consists of four main stages: movie chunking and preparation, pre-annotation design, pilot studies, and large-scale annotation. The process begins with movie and script selection, followed by scene segmentation and filtering of irrelevant or illicit content. Subsequently, annotation features and guidelines are defined and integrated into a web-based annotation tool. Pilot studies are then conducted to refine the annotation schema, train annotators, and improve consistency before scaling to large-scale human and LLM/MLLM-assisted annotation. Finally, the collected annotations undergo data cleaning, normalization, and analysis.

\subsection{Movie Selection}
Beyond general suitability, we applied a deliberate selection criterion: all chosen films depict recognizable, everyday social situations grounded in human reality. We excluded science-fiction, fantasy, and heavily action-oriented titles in favor of works that emphasize interpersonal relationships, social tensions, and moral dilemmas that audiences across cultures can recognize and relate to. To guide this selection, we consulted native speakers of each target language, asking them to recommend films known for naturalistic dialogue and a strong social or relational focus. We also limited our pool of movies to those available on YouTube with open access to their scenarios. 
\subsection{Scene Segmentation and Dialogue Extraction}
In the first stage, we transform unstructured screenplay text into a structured representation suitable for downstream annotation. Each raw screenplay is segmented into scenes using regular expressions that identify standard scene markers (e.g., \texttt{INT.}, \texttt{EXT.}, or numbered headings). Each scene is then submitted to an LLM (Gemini-2.5-Pro \footnote{https://docs.cloud.google.com/gemini-enterprise-agent-platform/models/gemini/2-5-pro}), which performs three tasks:\\
\textbf{Dialogue extraction}: the model identifies speaker turns and formats each utterance as \texttt{<Speaker: utterance>}. \\
\textbf{Scene details}: the model produces a concise description of the events, characters, and location of the current scene, which are well-defined in the screenplay.\\
\textbf{Movie summary till now}: we prompt the model to summarize all previous scenes until the current one by keeping important events and characters. This summary becomes a memory and helps annotators remember important events.\\

After formatting the scenes, we perform manual \textit{Scene Filtering and Verification}. We retain scenes relevant to the task and discard transition scenes, scenes with no meaningful events or dialog, and scenes containing explicit content. We also identify scene timestamps and manually verify the extracted dialog against the original screenplay, correcting inconsistencies in speaker attribution, dialog segmentation, and scene mapping. Human verification was conducted iteratively throughout pipeline development to identify recurring extraction errors and refine the preprocessing prompts accordingly. The verified dialogs are then mapped to their corresponding scene and cumulative summaries for downstream annotation.

\subsection{Pre-annotation}
After identifying the initial set of features to annotate, we developed a custom annotation tool along with an annotation guidelines document. The tool follows Human-Computer Interaction (HCI) best practices, including logical grouping of interface elements \cite{stureborg2023interface}, minimizing cross-page navigation \cite{perry-2021-lighttag}, and reducing typing burden by offering selectable options rather than free-text fields \cite{pei2022potato,hollender2010integrating}. The annotation guidelines document provides detailed annotation instructions and interface navigation. Further details about the annotation tool and guidelines are provided in Annotation Guidelines \footnote{\url{https://github.com/mbzuai-nlp/Social-Power-in-Dialogs}}.

\subsection{Pilot Study}
\textbf{Pilot 1.}
The first pilot involved three internal annotators and focused on assessing the clarity and feasibility of the annotation task. Annotators worked exclusively with textual inputs, dialogue turns, and a cumulative scene-level summary to label the proposed feature set. The goal was to surface ambiguous or poorly defined features, measure initial inter-annotator agreement, and identify which aspects of the schema required refinement before scaling up.\\
\textbf{Pilot 2.}
Building on the findings of Pilot 1, the second pilot extended the task to a multimodal setting: each scene was linked to its corresponding clip on YouTube, allowing annotators to draw on visual and acoustic cues alongside the screenplay text. In parallel, the annotation guidelines were substantially revised, condensed for readability, supplemented with worked examples, and restructured for easier navigation across sections. This iteration aimed to reduce annotator cognitive load, improve cross-annotator consistency, and evaluate whether access to video led to higher agreement or greater annotation confidence.
Across both pilot phases, schema refinement, guideline development, and pilot annotation iterations were conducted over seven months.
\subsection{Large Scale Annotation}
We conducted a human annotation campaign with native-speaker annotators for each target language.
Annotators received the annotation guidelines in advance and attended a practical training session in which each feature was explained in detail, ambiguous cases were discussed, and the interface was demonstrated. Annotators were recruited and paid through Prolific \footnote{https://www.prolific.com/}.
The large-scale annotation phase was completed over approximately three weeks following the finalization of the annotation schema and pilot studies. 
We provide further details on recruitment criteria and payment in Appendix \ref{annot_recruit} (Annotators recruitment).
\paragraph{Annotation statistics.}
Table~\ref{tab:stats} summarizes the corpus statistics. An \textit{annotated item} corresponds to an individual feature annotation. The difference in scene counts results from the filtering process in Section~4.2, with more Arabic scenes satisfying our selection criteria. Despite having fewer scenes and utterances, the French subset contains more annotated items because its selected scenes involve more interacting characters, generating more speaker profiles and directed speaker-dynamics edges.
\begin{table}[h]
\centering
\resizebox{\columnwidth}{!}{
\begin{tabular}{lcccc}
\hline
Language  & Scenes & Utterances & Annotated Items  \\
\hline
French  & 38 & 824  & 7998  \\
Arabic  & 62 & 1027 & 7838  \\
\hline
\end{tabular}
}
\caption{Corpus statistics}
\label{tab:stats}
\vspace{-0.8em}
\end{table}

\input{Chapters/annotation_schema}
\input{Chapters/experiments}

\section{Discussion and Conclusion}
We introduced a theoretically grounded, multilingual framework for studying social power in naturalistic dialog, comprising a social-science grounded annotation schema, a native-speaker annotation pipeline, and a custom interface designed for cross-lingual extensibility. By positioning social power as the primary object of annotation rather than a downstream proxy, and by capturing relational features directionally from both annotator and character perspectives, the framework enables analysis that prior computational work on power has not supported, including theory-of-mind reasoning, cross-cultural comparison, and fine-grained study of how distinct power bases are perceived and contested. Applying the framework to French and Egyptian Arabic movie dialogues, we find that annotators show strong agreement on observable demographic and contextual variables, whereas more interpretive dimensions, such as power asymmetry and intention alignment, are considerably more contested. Our evaluations further show that although current LLMs handle explicit surface-level social cues reasonably well, they continue to struggle with deeper socially grounded reasoning, especially when cultural understanding and theory-of-mind inference are involved. More broadly, these results contribute to ongoing efforts in computational social science that employ LLMs as proxies for human social behavior and interaction. While such models can reproduce certain patterns of social reasoning, they remain limited in capturing nuanced and culturally embedded power dynamics. This positions social power as a valuable diagnostic for evaluating cross-cultural social reasoning in AI systems, and motivates broader instantiations of the framework across languages, cultures, and modalities.

\section{Limitations}
This work represents an initial step toward multilingual and cross-cultural modeling of social power, but several limitations remain. First, the current dataset is limited to two languages, French and Egyptian Arabic, which restricts the generalizability of the findings across broader cultural settings. Therefore, we do not claim or assume that certain power types are associated with the cultures or draw any conclusions in this regard. A larger-scale release covering additional languages and substantially more dialogues is planned as a follow-up to this work, and will enable stronger statistical claims about cross-cultural patterns.

Second, while annotators were native speakers of each target language and culturally familiar with the depicted contexts, social power is ultimately a perceptual and interpretive construct. Two annotators per scene, although standard practice, cannot fully capture the range of plausible interpretations of contested dimensions such as power asymmetry or intention alignment. A larger annotator pool per item would allow a deeper study of perceptual variation and its demographic correlates.

Third, our benchmarking covers six state-of-the-art models, but the LLM and multimodal landscape evolve rapidly, and the specific numerical results reported here will date quickly. We mitigate this by releasing the schema, prompts, and annotated data so that future systems can be evaluated under identical conditions. Additionally, our multimodal evaluation is constrained by safety-aligned refusal and structural output failures in some systems, which reduce effective coverage and should be considered when interpreting agreement scores for those models.

\section{Ethical Considerations}
This work studies socially sensitive attributes and interpersonal power relations, including demographic characteristics, social class, religion, and hierarchy. To mitigate risks of harmful stereotyping or demographic overgeneralization, annotations were performed by trained native speakers familiar with the relevant cultural contexts, and the annotation guidelines explicitly encouraged annotators to avoid unsupported assumptions. Nevertheless, both human annotators and LLMs may still reproduce cultural biases present in media representations and societal norms.

The dataset is constructed from publicly available movie scripts and associated film content intended for public consumption. Explicit or harmful scenes were filtered during pre-processing where appropriate. Annotators were compensated for their work and informed in advance that some scenes could contain sensitive or emotionally difficult material. Finally, we emphasize that the framework is intended for research on social reasoning and cross-cultural analysis, not for profiling, surveillance, or automated judgment of individuals or social groups.
\bibliography{custom}
\bibliographystyle{acl_natbib}
\end{CJK}
\newpage
\appendix
\input{Chapters/appendixA}

\input{Chapters/appendixB}
\end{document}

%% file: Chapters/relatedwork.tex
\section{Social Power: A Primer}
Social power is one of the most contested concepts in social and political theory \cite{avelino2021theories}, yet scholars across multiple traditions have sought to define it. Marxist accounts root power in control over the means of production, whereby dominant classes shape the state, law, and ideology to reproduce class relations \cite{althusser2024ideology}. Weber defines it more broadly as the probability of carrying out one’s own will within a social relationship, even against resistance \cite{weber1978economy} a formulation that later inspired elite and pluralist theories. Against such zero-sum conceptions, Magee et al. \cite{magee20088} argue that power need not operate through domination or resistance; it can equally facilitate cooperation and coordination. Foucault departs further, treating power as relational and diffuse, and embedded in everyday practices rather than concentrated in any class \cite{foucault2012discipline}.
Although definitions of power remain contested, Raven et al. \cite{670675b5-0cdf-3cfa-899a-622e3191f2e6} provided one of the most systematic frameworks for understanding social influence by identifying several foundational bases of power. Their model originally proposed five key forms of power: (1) coercive power, which relies on force or threats; (2) reward power, exercised through incentives and benefits; (3) legitimate power, derived from socially accepted authority and norms; (4) expert power, grounded in knowledge and expertise; and (5) referent power, which stems from admiration, identification, or loyalty. Raven later extended this framework by introducing informational power, defined as influence achieved through the control and dissemination of information.
Scholars further broadened the concept of power. Mann \cite{mann2012sources} introduced the notion of ideological power, focusing on the capacity to shape values, beliefs, and perceptions,while Bourdieu's symbolic power \cite{bourdieu2018distinction} captured how power works through naming and classification to confer status and legitimacy.

\textbf{Power in computational approaches.}
In computational social science, power has been studied through the lens of language, network structure, and interaction patterns. \cite{danescu2012echoes} studied how people mirror each other's language and found that people with less power tend to mirror the language patterns of people with more power. \cite{danescu2013computational} built a computational model to score politeness using a large-scale corpus of Wikipedia and Stack Exchange requests, showing that powerful individuals tend to be less polite. \cite{diesner2005exploration} used network analysis to identify executives from email data, finding distinct network signatures associated with high-power positions.

Other work has modeled power more directly, primarily in organizational settings. \cite{bramsen-etal-2011-extracting} inferred hierarchical power relations from linguistic patterns in Enron emails, while \cite{prabhakaran-rambow-2013-written} studied four forms of power (hierarchical, situational, influence, and power over communication) and their manifestations in written dialog.
    
However, despite these contributions, two significant gaps remain. First, computational studies have predominantly examined power through linguistic or structural proxies, or within constrained organizational hierarchies, rather than modeling it as a multidimensional social phenomenon situated in broader interpersonal contexts. Second, the majority of existing studies rely on English-language data from Western settings, offering limited insight into how power dynamics operate across cultures and languages. Our framework addresses these limitations by modeling social power through theoretically grounded power types alongside demographic, relational, contextual, and perspective-dependent features, and by extending its study to multilingual, culturally situated interactions.
\begin{figure*}[hptb]
    \centering
    \includegraphics[width=\textwidth,height=6.5cm]{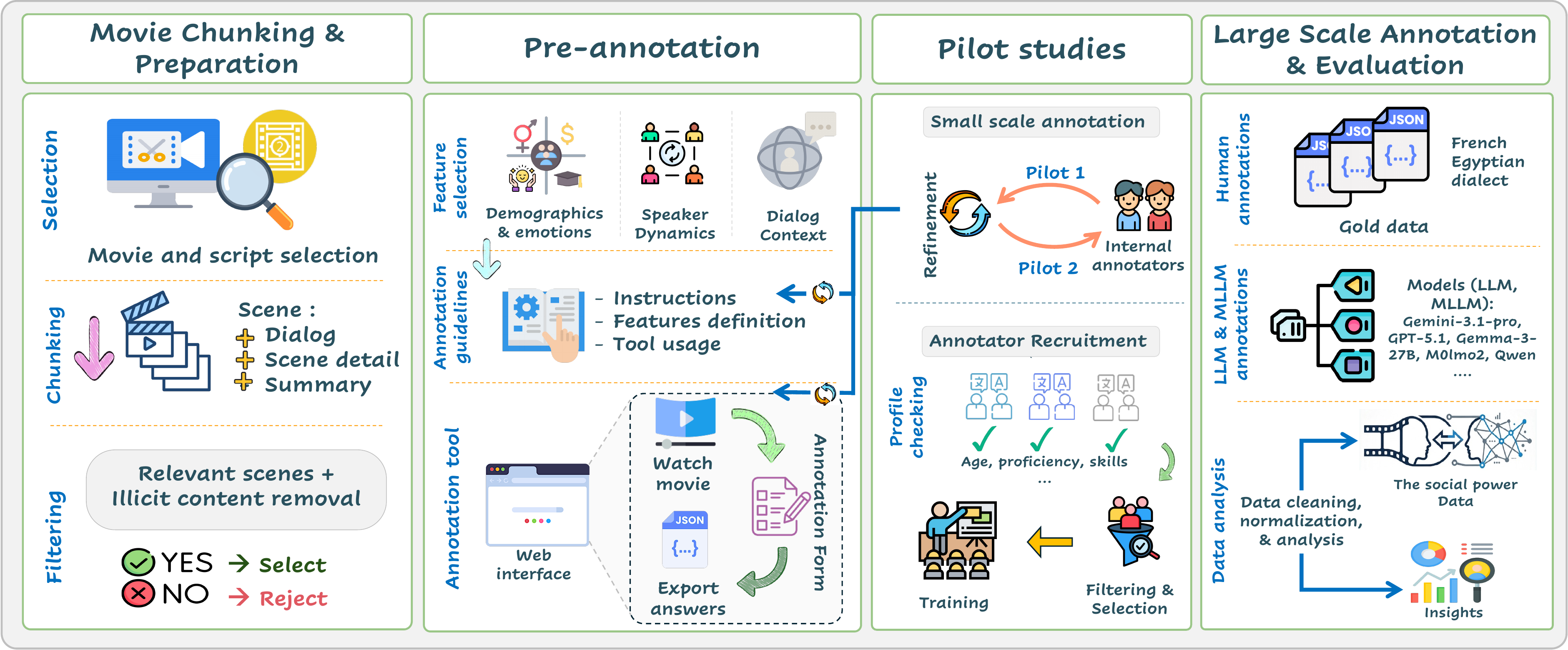}
    \caption{Overview of the annotation pipeline for movie dialogue dataset construction. The workflow comprises four stages: movie chunking and filtering, pre-annotation design (features, tool, and guidelines), pilot studies for annotator training and schema refinement, and large-scale human and LLM/MLLM annotation, followed by cleaning and analysis.}
    \label{fig:pipeline}
    
\end{figure*}

%% file: Chapters/principles.tex
\section{Annotation Framework}

\subsection{Movies as a Source of Social Power}
Movie screenplays have emerged as a valuable resource for studying social phenomena in language \cite{singh-etal-2022-hollywood}. Unlike synthetic or domain-limited corpora \cite{li2023diplomat}, screenplays offer human-crafted, naturalistic dialogue grounded in believable social contexts. They follow standardized formatting conventions that facilitate systematic extraction, and their longitudinal narrative structure allows characters to be traced across interactions \cite{batty2017screenwriting}, which helps capture how identity, relationships, and power evolve over time. Crucially, films are cultural artifacts; the language and social dynamics of characters reflect the values and interpersonal norms of their cultural context \cite{adilazuarda2024towards}, making screenplays particularly suited to cross-cultural analysis of social power.

\subsection{Design Principles}

The annotation schema and tooling were developed around six guiding principles.\\
\textbf{Predictive of power.}
Every feature in the schema was selected for its theoretical or empirical connection to social power. Social class, occupation, age, gender, relationship type, and goal alignment are all established correlates of power asymmetries in social interaction. Including them together allows the dataset to support both interpretive analysis and predictive modeling of power dynamics.\\
\textbf{Annotatable by humans.}
The schema had to be operationalizable by human annotators with reasonable reliability. This required translating abstract social constructs into observable, dialogue-grounded cues and choosing granularities for categories that balance expressiveness with cognitive tractability. Features that consistently produced low inter-annotator agreement during the pilot were revised or removed.\\
\textbf{Assessable by LLMs.}
In parallel with human annotation, the schema was designed to serve as a benchmark for large language models. Each feature is defined precisely enough to be prompted zero-shot or few-shot, enabling systematic comparison of model and human performance on the same task.\\
\textbf{Extendable to different languages and cultures.}
All features were designed to be language-agnostic and culturally broad. Where universal categories are insufficient, for example, the caste system in South Asian contexts, the schema provides culturally specific alternatives. This ensures the framework can be extended to new languages without structural redesign.\\
\textbf{Context and chronology preservation.}
Power dynamics are relational and cumulative: they cannot be reliably inferred from isolated utterances. The annotation pipeline, therefore, preserves narrative context by running scene summaries and character profiles that evolve throughout the film. \\
\textbf{Simplicity for annotators.}
Both the schema and the interface should be accessible to any annotator reading the guidelines or navigating the tool, without requiring a specialist background. Feature definitions should be concrete and dialogue-grounded, category sets bounded, and the interface should provide a seamless experience that minimizes cognitive load and allows annotators to focus on the judgment task itself. 
\subsection{Annotation schema}
\label{annot_schema}
We annotate five groups of features: demographic attributes of individual speakers, emotional states, power types, dynamic features characterizing the interaction between speakers, and dialogue context. The following subsections describe each feature and its category scheme. Full features and label inventories are provided in Appendix \ref{annot_schema_detail}.\\
\textbf{Demographic features.} We annotate demographic attributes that are theoretically associated with social hierarchy and interpersonal influence, including age group, gender, occupation, educational background, religion, ethnicity, social class, socio-economic status, marital status, and country of origin. These features provide a socially grounded context for understanding how power is expressed and perceived across cultures. \\
\textbf{Emotional State.} Social power has also been widely studied in psychology, particularly in relation to the psychological effects of power and social distance \cite{MAGEE202033}, as well as traits such as perspective-taking \cite{galinsky2016power}, competitiveness and advice-taking \cite{tost2012power}, and abstract thinking \cite{magee2013social}. To capture emotional dimensions, we adopt the six-level emotion taxonomy proposed by \cite{shaver1987emotion}.\\
\textbf{Power Types.} We adopt the extended taxonomy of social power introduced by \cite{670675b5-0cdf-3cfa-899a-622e3191f2e6}, and further expand it by including Ideological power.\\
\textbf{Speaker Dynamics.} Speaker dynamics focus on bilateral relations between interlocutors. For each interaction, we create two directed edges, $Speaker A \rightarrow Speaker B$ and $Speaker B \rightarrow Speaker A$, since these relationships may be asymmetric. We annotate the following features: perceived power asymmetry, social-status difference, familiarity, relationship type, and goal alignment. Together, these variables capture interaction-level manifestations of power and support the analysis of socially situated reasoning and theory-of-mind inference.\\
\textbf{dialogue Context.} We annotate contextual properties of the interaction environment, including whether the dialogue occurs in personal or professional settings and whether the interaction takes place in public or private spaces.

%% file: Chapters/experiments.tex

\begin{table*}[t]
\centering
\scriptsize
\setlength{\tabcolsep}{3pt}
\renewcommand{\arraystretch}{1.3}
\begin{tabular}{llccccccccccccccccccccc}
\toprule
\textbf{Lang.} & \textbf{Evaluator pair} &
\rotatebox{90}{Age group} &
\rotatebox{90}{Gender} &
\rotatebox{90}{Marital status} &
\rotatebox{90}{Religion} &
\rotatebox{90}{Educ. tier} &
\rotatebox{90}{Occup. tier} &
\rotatebox{90}{SEC} &
\rotatebox{90}{Social class} &
\rotatebox{90}{Emotions} &
\rotatebox{90}{Rel. category} &
\rotatebox{90}{Relationship} &
\rotatebox{90}{Familiarity} &
\rotatebox{90}{Intent align.} &
\rotatebox{90}{Power diff.} &
\rotatebox{90}{Power diff. persp.} &
\rotatebox{90}{Powers} &
\rotatebox{90}{Social diff.} &
\rotatebox{90}{Social diff. persp.} &
\rotatebox{90}{Intent persp.} &
\rotatebox{90}{Domain} &
\rotatebox{90}{Privacy} \\
\midrule

\multirow{7}{*}{\rotatebox{90}{French}}
& Gemini-3.1-Pro           & 0.57 & 0.86 & 0.46 & \textbf{1.00} & 0.98 & 0.54 & 0.58 & 0.57 & 0.54 & 0.82 & 0.74 & 0.74 & \textbf{0.70} & 0.32 & 0.32 & 0.50 & 0.40 & 0.41 & 0.61 & 0.47 & 0.66 \\
& GPT-5.1                  & 0.49 & \textbf{0.98} & 0.39 & \textbf{1.00} & 0.93 & 0.54 & 0.40 & 0.38 & 0.54 & 0.46 & 0.63 & 0.70 & 0.62 & 0.34 & 0.34 & 0.29 & 0.12 & 0.15 & 0.32 & 0.48 & 0.24 \\
& Qwen-2.5-14B $\diamond$            & 0.55 & 0.97 & 0.38 & 1.00 & 0.61 & 0.55 & 0.32 & 0.31 & 0.35 & 0.43 & 0.52 & 0.53 & 0.42 & 0.20 & 0.18 & 0.33 & 0.17 & 0.17 & 0.38 & 0.22 & 0.40 \\
& Gemma-3-27B $\diamond$             & 0.59 & 0.71 & 0.37 & 1.00 & 0.76 & 0.51 & 0.49 & 0.48 & 0.48 & 0.53 & 0.62 & 0.53 & 0.38 & 0.35 & 0.21 & 0.25 & 0.27 & 0.21 & 0.06 & 0.53 & 0.45 \\
& Molmo2*$\diamond$          & 0.50 & 0.77 & 0.23 & \textbf{1.00} & 0.72 & 0.50 & 0.35 & 0.34 & 0.33 & 0.41 & 0.50 & 0.61 & 0.34 & 0.12 & 0.18 & 0.32 & 0.09 & -0.02 & 0.40 & 0.52 & 0.56 \\
& Gemini-3.1-Pro*$\diamond$  & 0.74 & 0.98 & 0.36 & 1.00 & 0.96 & 0.65 & 0.56 & 0.56 & 0.54 & 0.84 & 0.77 & 0.80 & 0.69 & 0.33 & 0.37 & 0.62 & 0.44 & 0.51 & 0.52 & 0.52 & 0.68 \\
\rowcolor{violet!20!blue!10}  & Human                     & \textbf{0.78} & \textbf{0.98} & \textbf{0.64} & \textbf{1.00} & \textbf{1.00} & \textbf{0.85} & \textbf{0.80} & \textbf{0.78} & \textbf{0.61} & \textbf{0.85} & \textbf{0.78} & 0.76 & 0.65 & \textbf{0.57} & \textbf{0.57} & 0.59 & \textbf{0.53} & \textbf{0.57} & \textbf{0.94} & \textbf{0.90} & \textbf{0.81} \\
\midrule

\multirow{7}{*}{\rotatebox{90}{Arabic-Eg.}}
& Gemini-3.1-Pro           & 0.53 & \textbf{1.00} & 0.47 & 0.53 & 0.60 & 0.63 & 0.81 & 0.81 & 0.43 & 0.92 & 0.83 & 0.77 & 0.57 & 0.23 & 0.28 & 0.36 & 0.71 & 0.64 & 0.49 & 0.71 & 0.58 \\
& GPT-5.1                  & 0.71 & \textbf{1.00} & 0.53 & 0.24 & \textbf{0.67} & 0.58 & 0.57 & 0.55 & 0.47 & 0.85 & 0.83 & 0.73 & 0.48 & 0.12 & 0.19 & 0.44 & 0.55 & 0.49 & 0.17 & 0.69 & 0.58 \\
& Qwen-2.5-14B$\diamond$           & 0.47 & 0.99 & 0.33 & 0.19 & 0.64 & 0.60 & 0.51 & 0.51 & 0.33 & 0.81 & 0.73 & 0.66 & 0.44 & 0.08 & 0.15 & 0.37 & 0.40 & 0.39 & 0.08 & 0.43 & 0.33 \\
& Gemma-3-27B$\diamond$              & 0.72 & 0.98 & 0.39 & 0.40 & 0.46 & 0.53 & 0.71 & 0.68 & 0.40 & 0.81 & 0.71 & 0.74 & 0.32 & 0.14 & 0.14 & 0.14 & 0.51 & 0.40 & 0.33 & 0.58 & 0.45 \\
& Molmo2*$\diamond$          & 0.43 & 0.97 & 0.39 & 0.34 & 0.39 & 0.40 & 0.67 & 0.65 & 0.33 & 0.04 & 0.39 & 0.42 & 0.03 & 0.17 & -0.24 & 0.51 & 0.44 & -0.22 & 0.27 & 0.55 & 0.37 \\
& Gemini-3.1-Pro*$\diamond$   & \textbf{0.81} & 1.00 & 0.52 & 0.41 & 0.61 & 0.56 & 0.91 & 0.89 & 0.45 & 0.92 & 0.83 & 0.74 & 0.59 & 0.20 & 0.19 & 0.41 & 0.64 & 0.60 & 0.39 & 0.73 & 0.68 \\
\rowcolor{violet!20!blue!10} & Human  & 0.62 & \textbf{1.00} & \textbf{0.86} & \textbf{0.85} & 0.52 & \textbf{0.77} & \textbf{0.93} & \textbf{0.90} & \textbf{0.68} & \textbf{0.97} & \textbf{0.85} & \textbf{0.83} & \textbf{0.69} & \textbf{0.44} & \textbf{0.38} & \textbf{0.59} & \textbf{0.83} & \textbf{0.78} & \textbf{0.93} & \textbf{0.89} & \textbf{0.68} \\
\bottomrule
\end{tabular}
\caption{Inter-annotator agreement scores across evaluator pairs and dimensions for French and Egyptian Arabic datasets.(*) refer to Multimodal LLM's; ($\diamond$) indicates models with incomplete coverage, whose agreement scores are computed only over successfully annotated items. Ethnicity and country features were discarded due to perfect agreement across all.}
\label{tab:iaa-power}
\vspace{-0.5em}
\end{table*}
\vspace{-0.5em}
\section{Evaluation}
\label{sec:experiments}

We conduct two lines of evaluation: i) an inter-annotator agreement (IAA) study to validate the quality and reliability of the human annotations, and ii) a benchmarking study comparing six models against the human annotations across all annotation tiers.

\subsection{Inter-Annotator Agreement}
\label{sec:iaa}

\paragraph{Setup.}
All movies were annotated by two independent native-speaker annotators, yielding parallel annotations for every task. We compute agreement separately for each feature and report Gwet's AC2 \cite{gwet2014handbook} for categorical and ordinal features; a statistic that is more robust than Cohen's $\kappa$ under skewed distributions \cite{wongpakaran2013comparison}. For multi-label features such as power types and emotions, we report the mean Jaccard similarity over annotation pairs, as each annotator may assign any subset of labels.
\subsection{LLM and MLLM Benchmarking}
We benchmark a range of state-of-the-art LLMs and MLLMs. For text-only LLMs, we evaluate two closed-weight models, Gemini-3.1-Pro \footnote{https://deepmind.google/models/gemini/pro/} and GPT-5.1 \footnote{https://openai.com/index/gpt-5-1/}, and two open-weight models, Gemma-3-27b \footnote{https://deepmind.google/models/gemma/gemma-3/} and Qwen-2.5-14B \footnote{https://qwen.ai/blog?id=qwen2.5}. For multimodal evaluation, we use Gemini-3.1-Pro with video input \footnote{https://deepmind.google/models/gemini/pro/} and Molmo2 \cite{clark2026molmo2}, allowing us to assess whether access to visual context improves power identification beyond what is available from dialogues alone. Each model receives the dialog, the scene summary, the running summary of prior scenes, and the full annotation schema, including feature definitions and permitted value sets (see Appendix \ref{prompts} for the prompts). We used an NVIDIA RTX 6000 PRO machine to run the experiments. As for GPT-5.1 and Gemini models, the cost is estimated at approximately 100 USD each. We also fixed the temperature of the models to 0.1. After that, we performed cleaning and normalization steps, including stripping special characters from speaker names, removing incomplete annotations such as missing edges in speaker dynamics, and discarding hallucinated profiles. Hallucinated profiles were defined as generated profiles or speaker-dynamics edges for characters absent from the scene dialog and were automatically detected by matching generated names against the scene speaker list. Profiles with no matching speaker or with the speaker name set to ``NA'' were discarded, as they could not be reliably mapped to a character.
\begin{figure*}[t]
    \centering
    \includegraphics[width=\textwidth,height=5.5cm]{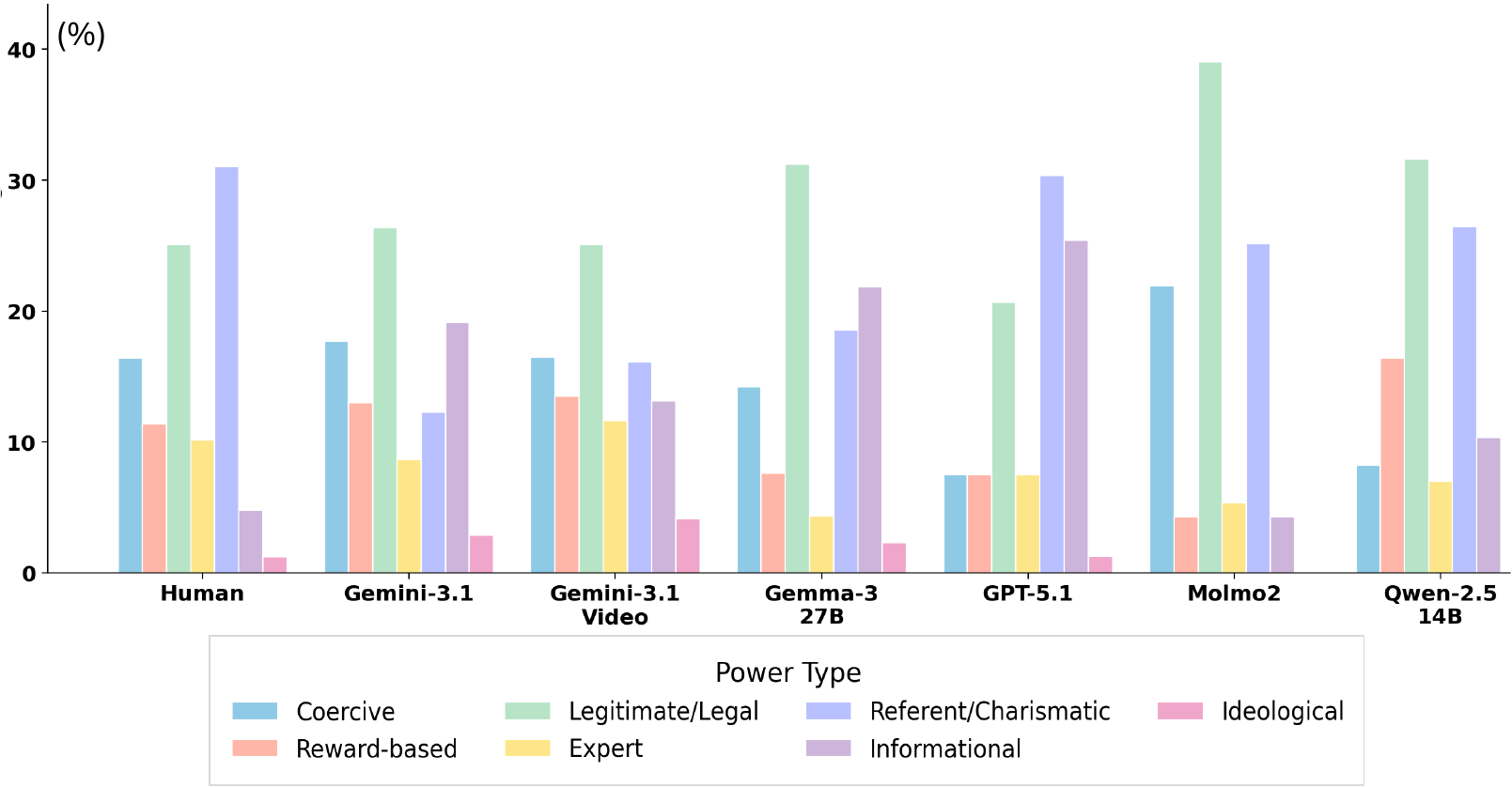}
    \caption{Power type distribution across models and human annotators}
    \label{fig:power_dist}
    \vspace{-1em}
\end{figure*}
\section{Analysis}
\subsection{Human and LLM Comparison}

\paragraph{Overall patterns.}
Table~\ref{tab:iaa-power} reports agreement between human annotators and six models across the French and Arabic-Egyptian subsets. Agreement varies substantially by feature type. Among the text-only models, performance is strongest on relatively observable attributes such as gender, relationship category, familiarity, and educational tier. Gender agreement approaches or reaches 1.00 across most systems, while relationship-related variables also achieve comparatively high agreement. These results suggest that models perform better when social cues are explicit or directly recoverable from the dialog and its context.

Human annotators similarly agree most on demographic and contextual variables. In the Arabic-Egyptian subset, agreement is particularly high for socio-economic class (0.93), social class (0.90), relationship category (0.97), and familiarity (0.83). By contrast, agreement is substantially lower for power difference, social-status asymmetry, and intention alignment, which require relational inference and perspective-taking. Such disagreement may therefore reflect genuine variation in perceptions of social power rather than annotation noise.

Human--model comparisons reinforce this distinction. Models approach human agreement on several demographic and contextual attributes but perform substantially worse on emotions, speaker dynamics, and socially interpretive variables. In Egyptian Arabic, religion agreement ranges from 0.19 to 0.53 among text models, largely because models infer religious identity from culturally common expressions such as \textit{Inshallah} (God willing) or references to Allah (God), whereas annotators often assign ``NA'' without sufficient evidence. This illustrates how culturally shared linguistic cues can be overgeneralized into demographic attributes. Similarly, GPT-5.1 achieves only 0.12--0.15 agreement on social-status difference in French. Overall, the largest human--model gaps occur on dimensions involving implicit power structures, perspective-taking, and culturally grounded social inference.
\paragraph{Model Coverage.}
Coverage denotes the proportion of gold items receiving complete, usable annotations. GPT-5.1 and Gemini-3.1-Pro achieve 100\% coverage, followed by Gemma-3-27B (92.4\%) and Qwen-2.5-14B (89.9\%), while both multimodal models cover only 55.7\%. Gemini-3.1-Pro (video) frequently skips scenes involving children, likely due to safety refusals, while Molmo2 produces incomplete outputs and unreliable speaker identification. Thus, multimodal agreement scores in Table~\ref{tab:iaa-power} reflect only successfully annotated items and cannot establish a general benefit from multimodal input.

\subsection{Which Power Types Are Most Contested?}
Qualitative analysis shows that Referent/Charismatic power is the most contested category, most often in relation to the absence of any identifiable power label. Annotators disagreed on whether Referent power was present in 23\% of such cases, compared to only 9\% for unidentified-versus-Legitimate distinctions. This suggests that personal influence is less consistently perceived than institutional authority.
Among disagreements involving two explicit labels (13\% of cases), three recurring patterns emerge. First, Expert versus Referent/Charismatic disagreements (4\%) appear only in the Arabic subset, where knowledgeable and charismatic authority often overlap. Second, Legitimate versus Referent disagreements (4\%) reflect tension between institutional and personal influence. Third, the most theoretically significant disagreements involve Coercive versus Legitimate power, concentrated in a film centered on an abusive father. One annotator interpreted the character's behavior as paternal authority, while another viewed it as domination and coercion. Rather than reflecting annotation error, these cases reveal genuine conceptual ambiguity at the boundary between legitimate and coercive authority.
\begin{figure}[hptb]
    \centering
\includegraphics[width=0.49\textwidth,height=7.5cm]{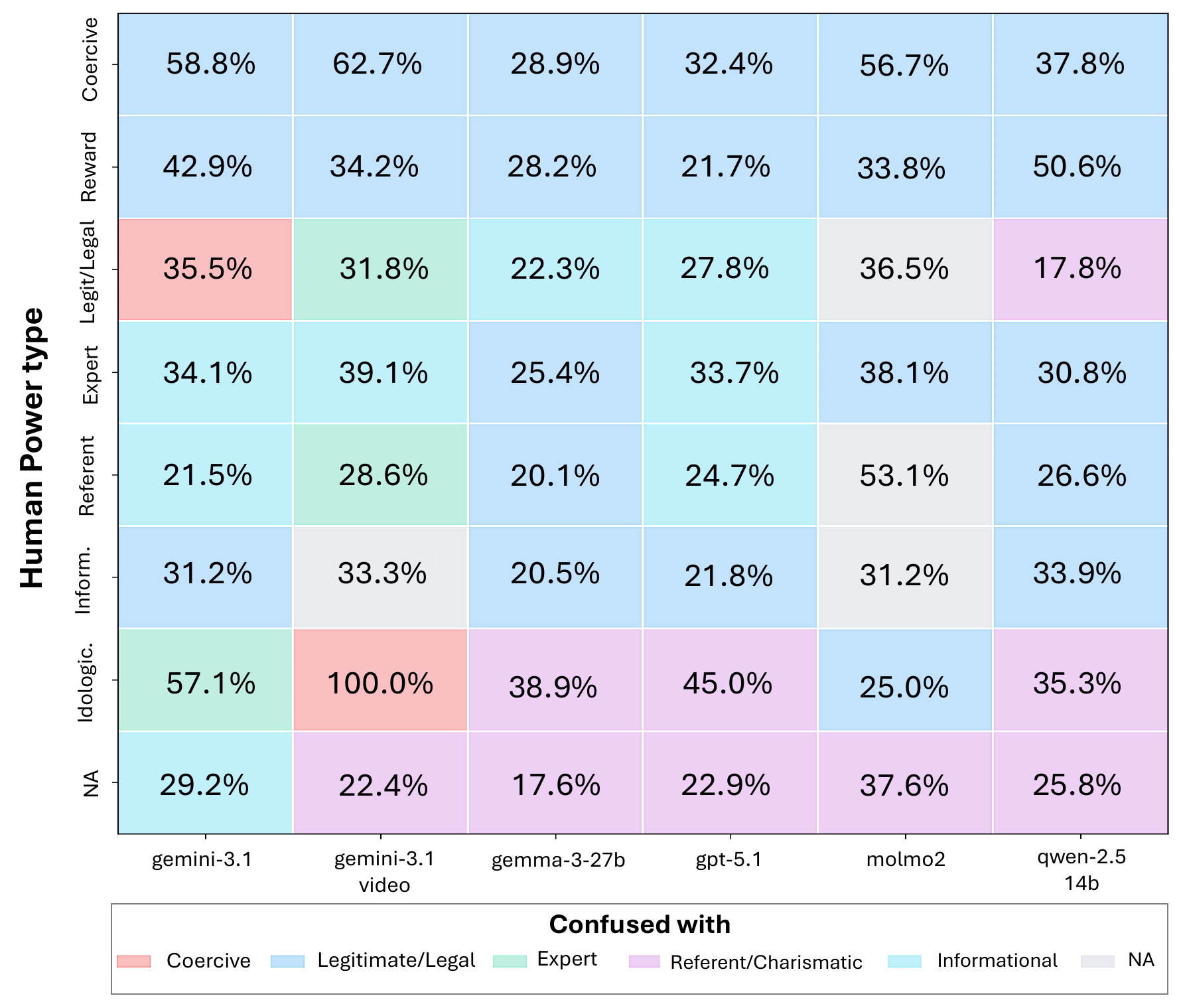}
    \caption{Most frequent human–model power-type confusions across models over the French and Egyptian Arabic data.}
    \label{fig:error_analysis}
  \vspace{-1em}
\end{figure}
\paragraph{Power-Type Confusion Analysis.}

Figure \ref{fig:error_analysis} shows the dominant human–model power-type confusions, where each cell shows the model label most frequently assigned when it differs from the corresponding human power-type annotation. Across models, Coercive and Reward-based power are most frequently collapsed into Legitimate/Legal authority, suggesting that models tend to normalize forms of control into institutionally sanctioned authority. In contrast, Referent/Charismatic and Expert power exhibit greater instability, often being reinterpreted as Informational or Expert authority. Open-weight and multimodal LLMs additionally show stronger collapse into NA labels, particularly Molmo2, indicating difficulties in grounding subtle or socially implicit forms of power even with visual context. Overall, the results suggest that current models recover explicit and institutional forms of authority more reliably than relational or culturally situated forms of influence.

%% file: Chapters/appendixA.tex
\clearpage
\section{Appendix}

This appendix provides supplementary material supporting the proposed social power framework and annotation pipeline. We first present the detailed annotation schema and feature definitions, covering demographic attributes, emotional states, power types, speaker dynamics, and dialogue context. We then describe the annotator recruitment protocol, training process, and qualification procedure used during dataset construction. Finally, we provide the prompts employed throughout the framework, including the movie processing and scene extraction prompts, as well as the social power annotation prompts used during large-scale annotation.




\subsection{Detailed Annotation Scheme}
\label{annot_schema_detail}

\begin{figure}[h]
    \centering
    \includegraphics[width=0.4\textwidth,height=5cm]{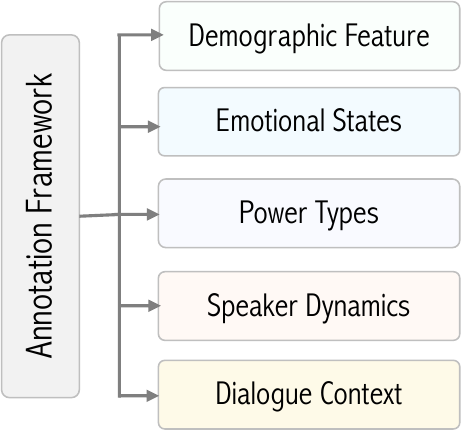}
    \caption{Overview of the proposed annotation framework and its five major components for modeling social power in dialogue.}
    \label{fig:annotation_framework}
\end{figure}

The annotation framework models social power as a multidimensional phenomenon expressed through demographic cues, interpersonal relations, contextual grounding, and socially situated reasoning. The schema combines observable attributes, such as age or occupation, with interpretive relational variables, including perceived power asymmetry and intention alignment. As illustrated in Figure~\ref{fig:annotation_framework}, the framework is organized into five interconnected components spanning demographic features, emotions, power types, speaker dynamics, and dialogue context.

\subsubsection{Demographic features}

\textbf{Age group.}
Following \cite{Warner_Meeker_Eells_1960} and \cite{skiadas2017quantitative}, we group age into five broad cohorts: \textit{Child, Teenager, Adult, Middle-aged, Elderly}. Annotators apply their own cultural judgment, as age boundaries vary across cultures.

\textbf{Gender.}
Gender is represented through \textit{Male}, \textit{Female}, \textit{Other}, and \textit{NA} categories. 

\textbf{Occupation.}
We follow the World Values Survey (WVS) coding \cite{haerpfer2022world}, distinguishing employed/self-employed roles (manager, professional, technical, clerical, service, agricultural, craft, operator, elementary, armed forces) from non-employed statuses (retired, student, housewife, unemployed).
Occupation is included because professional role and labor position frequently function as visible indicators of institutional authority, expertise, and socio-economic positioning within interaction.

\textbf{Educational Tier.}
Following ISCED \cite{schneider2022classification}, we group education into three coarse-grained adaptation of the International Standard Classification of Education (ISCED) \citep{schneider2022classification}.  The framework distinguishes \textit{Higher} (Doctoral, Master’s, Bachelor’s), \textit{Medium} (diploma, secondary), and \textit{Lower} (elementary or below). This coarser grouping reflects the practical difficulty of inferring fine-grained educational attainment from movies, where the fine-grained information is not always easily inferred.

\textbf{Marital status.}
Marital status categories are adapted from the World Values Survey coding \cite{}, namely: \textit{Married, Single, Divorced, Separated, Widowed, Living together as married} with the addition of \textit{In a relationship / Engaged} to accommodate culturally salient forms of partnership and engagement.

\textbf{Religion.}
Religion is represented through broad categories including \textit{Muslim}, \textit{Buddhist}, \textit{Christian}, \textit{Hindu},  \textit{Jewish}, \textit{Atheist}, \textit{Other}, and \textit{NA}.
The framework intentionally avoids sectarian distinctions unless they are central to the story where annotators can select "Other" to specify. 

\textbf{Ethnicity.}
We use a broad regional taxonomy namely:
\textit{African / African Descent; Arab / Middle Eastern / North African; Central Asian; South Asian; East Asian; Southeast Asian; Pacific Islander; European / White; Latino / Hispanic; Indigenous / Native Peoples; Mixed / Multi-ethnic; Other; NA.}

\textbf{Social Class.}
Following Bourdieu \cite{bourdieu2018distinction}, social class is inferred from observable cues such as occupation, speech style, residential setting, clothing, and naming conventions \cite{kraus2014sartorial, nagi2011stratification} and coded as \textit{upper}, \textit{middle}, or \textit{lower}. Annotators are instructed to apply their own cultural framing, as class markers vary across societies.

\textbf{Socio-economic Class.}
We distinguish social class from socio-economic class (SES), as the latter indexes an individual's position within a power hierarchy through relatively objective indicators of resources and capital, such as income, wealth, educational attainment, and occupational prestige \cite{diemer2013best}. The framework distinguishes \textit{Upper}, \textit{Middle}, and \textit{Lower} categories.

\textbf{Country of origin.}
The country from which a character originates and is distinct from the location where the dialogue takes place.

\subsubsection{Emotional state}
The framework adopts the hierarchical taxonomy proposed by \citet{shaver1987emotion}, which organises emotions around six primary categories: \textit{Love}, \textit{Joy}, \textit{Surprise}, \textit{Anger}, \textit{Sadness}, and \textit{Fear}.

Emotional states are annotated at the scene level rather than as stable character traits. The inclusion of affective information reflects the close relationship between emotion, persuasion, conflict escalation, social dominance, and interpersonal alignment.

\subsubsection{Power Types}
The framework operationalizes power using an extended adaptation of French and Raven’s taxonomy of social power \citep{670675b5-0cdf-3cfa-899a-622e3191f2e6}. The schema includes coercive, reward-based, legitimate, expert, referent, informational, and ideological forms of power.

This multidimensional operationalization reflects the view that social power cannot be reduced to institutional authority alone. Instead, power may emerge from expertise, social prestige, ideological influence, charisma, control of information, or the ability to impose sanctions.
The framework further assumes that multiple forms of power may coexist within a single interaction for a single character and may vary across cultural settings.

\subsubsection{Speakers Dynamics}
 
\textbf{Power difference.}
Power difference captures the relative relationship between interlocutors within a given interaction, namely: \textit{Higher}, \textit{Equal}, or \textit{Lower} power, with \textit{NA} reserved for unclear cases. The framework models this relation directionally because social influence is frequently asymmetric. Judgements are collected both from the annotator’s global interpretation and from the inferred perspective of individual interlocutors. 
This distinction allows the framework to capture divergences between annotators and character perception, thereby supporting the study of theory-of-mind reasoning in dialogues. 

\textbf{Social status difference.}
Social-status difference represents perceived asymmetries in prestige, hierarchy, or social standing between interlocutors. The framework distinguishes \textit{High}, \textit{Equal}, and \textit{Lower} social status. Similarly to power difference, judgements are collected both from the annotator’s perspective and the character's perspective.

\textbf{Familiarity}

Familiarity operationalises the degree of social closeness between interlocutors. The feature draws on traditions in sociolinguistics and politeness theory that treat social distance as a major determinant of interactional behaviour, accommodation, and communicative style. Two categories are considered \textit{Familiar} and \textit{Unfamiliar}.

\textbf{Interlocutor Goal Alignment.}
Intention alignment captures whether interlocutors pursue \textit{aligned}, \textit{complementary}, or \textit{conflicting} goals within the interaction.

The feature is motivated by recent work on conversational motives and social reasoning \citep{yeomans2022conversational}. Intention alignment is annotated both from the characters perspectives and from the annotator’s broader narrative perspective, enabling analysis of misunderstanding, deception, and socially situated inference.

\textbf{Relationship Category and Type.}
Following \cite{tigunova2021pride}, relationships are categorized into three groups: \textit{Family} (e.g., parent, spouse, sibling, child), \textit{Social} (e.g., friend, neighbour, enemy, fan), and \textit{Professional} (e.g., colleague, doctor--patient, teacher--student).
This feature captures structurally salient interpersonal ties that shape expectations surrounding authority, obligation, familiarity, and emotional expression.

\subsubsection{Dialogue Context} 
The framework additionally annotates contextual properties of the interaction environment. \textbf{Location domain} distinguishes between \textit{professional} and \textit{personal} settings, while \textbf{location privacy} captures whether the interaction occurs in \textit{public} or \textit{private} space.

These contextual variables are included because social behaviour and power expression are subject to situational environment, audience presence, and institutional setting.

\begin{figure*}[t]
\centering
\begin{tcolorbox}[
title={Movie preprocessing},
width=0.85\textwidth,
colback=white,
colframe=gray,
arc=0pt,
outer arc=5pt,
boxrule=0.5pt,
leftrule=2pt,
rightrule=2pt,
right=2pt,
left=2pt,
top=2pt,
bottom=2pt,
toprule=0pt,
bottomrule=2pt]
\fontfamily{pcr}\selectfont
\scriptsize

You are a movie script processing assistant. You are given a chunk of a movie script.Your task has three parts:\\

\textbf{1. Extract Dialogue} \\
    - Make sure to group dialogues that belong to the same scene, think step by step when grouping dialogues. \\
    - Only extract lines of actual dialogue. \\
    - Format: Each line must follow this exact format: ``CHARACTER: dialogue". \\
    - Ignore all descriptions, scene directions, or narration. \\
    - If the line contains any **illicit or harmful content** (e.g., self-harm, abuse, jailbreak), **replace that entire line** with: ``<ILLICIT>".\\
    - Do not merge dialogues that do not belong to the same scene. Split wisely and when necessary.\\
    - The scenes start usually with {examples} Therefore, split accordingly. \\
    - If there is no dialogue in a scene no need to include it. \\
--- \\
\textbf{2. Build Memory of Events and Characters} \\
    
    1. Create a summary of the current events from the provided script\\
    2. Consider the past events in the  \{memory\} \\
    3. Merge the two summaries to create a coherent memory that describes all important events that happened. A user who is not familiar with the movie should be familiar after reading the summary. \\
    - Each memory should provide a brief description of the persons involved in the scene like (Age, occupation, education, ...etc). \\
    - The summary should include Before events and Current events. In before describe major events in previous scenes. And now, what is currently happening in this scene. \\
    - Do NOT reset memory each time. Think of memory like a snowball: each group adds to it. \\
    - The final memory should feel like an **ongoing, concise summary of everything important so far** (including both past and new information).
    - The summary should be in the same language as the original script in \{language\}. \\
    - Refine the memory by incorporating new events and removing less relevant ones.\\ 
    - Make sure that the summary will help the reader understand the movie no matter from where they start reading it.\\
        
\textbf{3. Scene specific details} \\
    - For each dialogue include a  description of the people, the scene, and relevant information to understand the current dialogue.  \\

**Input Memory**:\\
    \{memory\}\\
        
**Script Chunk**:\\
    \{dialogues\} \\
        
---\\
\textbf{Output Format} \\
    - **dialogues**: a list of dialogue lines from that group, formatted as "CHARACTER: dialogue". \\
    - **scene details**: (A concise description of the current scene ) short around 200 words. \\
    - **overall summary**: (The comprehensive, cumulative summary of the entire story so far) around 400 words. \\ 
    - The output must be a JSON. 

\end{tcolorbox}
\end{figure*}

\subsection{Annotators recruitment}
\label{annot_recruit}
\vspace{-0.5em}
We recruited four annotators, two per language, through the Prolific platform, subject to the following criteria: (i) native speakers of the target language with strong familiarity with the associated cultural norms and conventions; (ii) aged 18 or older; (iii) holding at least a Bachelor's degree, with proficient reading and writing competence in English; and (iv) they must possess a Prolific account. All annotators completed a two-hour online training session, followed by two qualification tasks designed to assess their understanding of the schema before proceeding to the main annotation. Annotators were compensated at a rate of 12 USD per hour, with compensation also covering time spent watching the movies and reviewing the annotation guidelines.

\vspace{0.5em}

\subsection{Prompts}
\label{prompts}


In the following section, we present the ``Movie Processing Prompt" used during the movie pre-processing and scene extraction stages, followed by the ``Annotation Prompt" which was carefully designed to capture the nuanced aspects of social power during the annotation stage of the proposed framework.

%% file: Chapters/appendixB.tex
\begin{figure}[!hptb]
\centering
\begin{tcolorbox}[
title={Annotation Prompt},
width=0.5\textwidth,
colback=white,
colframe=gray,
arc=0pt,
outer arc=5pt,
boxrule=0.5pt,
leftrule=2pt,
rightrule=2pt,
right=2pt,
left=2pt,
top=2pt,
bottom=2pt,
toprule=0pt,
bottomrule=2pt]

\fontfamily{pcr}\selectfont
\scriptsize
You are a movie script processing assistant. You are given a scene from a movie script, including a dialogue, an overall summary and a scene summary. \\       
\textbf{**Disclaimer**:} Some of the scenes may contain sensitive content, arguments, explicit language. We expect you to be able to process such content and provide your analysis without any issue. You can ignore explicit content as well and focus on the analysis. 
This processing is intended for research purposes to analyze the representation of different social groups in movies and how they interact with each other. We are not asking you to judge the content but rather to analyze it based on the instructions provided. \\     
Your task has three parts, think step by step: \\
\textbf{1. Demographic features} \\ 
- For each character or person involved in the scene, provide the following: \\
    - \textbf{Name} \\
    - \textbf{Age group:} Child, teenage, adult, middle age, elderly \\
    - \textbf{Gender:} Male, female, other, NA \\
    - \textbf{Ethnicity:}  'African / African Descent', 'Arab / Middle Eastern / North African', 'Central Asian',' South Asian (e.g., Indian, Pakistani, Bangladeshi)', 'East Asian (e.g., Chinese, Japanese, Korean)', 'Southeast Asian (e.g., Filipino, Vietnamese, Thai)',' Pacific Islander / Oceanian', 'European / White', 'Latino / Hispanic', 'Indigenous / Native Peoples', 'Mixed / Multi-ethnic', 'Other','NA'.  \\  
    -\textbf{ Marital Status: }'Married','Single','Divorced', 'Separated', 'Widowed', 'Living together as married', 'In a relationship/engaged','NA' \\
    - Educational tier: low education ( Elementary or Secondary), medium education (High School, Diploma (technical or vocational)), high education (Bachelor’s, Master’s, Doctoral'),  NA (When the information is unclear) \\
    - \textbf{Religion:} 'Buddhist','Christian', 'Hindu', 'Muslim', 'Jew', 'Other religion','Atheist','NA' \\
    - Socio-economic class : Upper, middle, lower, NA if not clear or hard to determine. \\
    - \textbf{Social class:}  Upper, middle, lower, NA if not clear or hard to determine.\\
    -\textbf{ Occupation tier:} Employed/self-employed , No/unpaid, NA if hard to determine. \\ 
    - \textbf{Country:} Name of the country, NA when hard to infer, Fictional if the country is fictional. 
    - \textbf{Emotions:} Select one or many emotions from the following category:  \\
             Love, Joy, Anger, Sadness, Fear, Surprise.\\ 
    - \textbf{Power type:} Select one or many power types for this person: \\
         'Coercive', 'Reward-based', 'Legitimate/Legal', 'Expert','Referent/Charismatic', 'Informational', 'Ideological', or 'NA'. \\
        'Coercive': 'Coercive power relies on the ability to punish or enforce consequences for disobedience. E.g: a manager threatening to fire an employee', \\
        'Reward-based': 'Reward power comes from the ability to provide positive incentives or benefits. E.g: a manger offering bonus',\\
        'Legitimate/Legal': 'Legitimate power is based on a person’s formal position or authority. E.g: a police officer enforcing the law, a mother making rules for her children',\\
        'Expert': 'Expert power comes from possessing specialized knowledge, skills, or experience that others respect. E.g: a doctor, a teacher explaining a concept',\\
        'Referent/Charismatic': 'is based on personal qualities that make others admire or trust. E.g: a charismatic, a celebrity, a charismatic person',\\
        Informational': 'Informational power arises from controlling access to important data or insights. E.g: a journalist with exclusive information, a person with insider knowledge',\\
        'Ideological': 'Ideological power is rooted in shared beliefs or values that inspire others to act. E.g: a political leader rallying supporters, a religious leader guiding followers. \\
        'NA': 'hard to determine or the dialogue does not exhibit any type of power for a speaker.',\\    
        \textbf{2. Speakers Dynamics}\\
    - For each pair of persons you detected earlier. 
\end{tcolorbox}
\end{figure}

\begin{figure}[!hptb]
\centering
\begin{tcolorbox}[
title={Annotation Prompt Continued},
width=0.5\textwidth,
colback=white,
colframe=gray,
arc=0pt,
outer arc=5pt,
boxrule=0.5pt,
leftrule=2pt,
rightrule=2pt,
right=2pt,
left=2pt,
top=2pt,
bottom=2pt,
toprule=0pt,
bottomrule=2pt]

\fontfamily{pcr}\selectfont
\scriptsize 
provide the following: \\
   \textbf{ 1. From person A to B:} \\
        - \textbf{Relationship Category:} Family, social, professional, Other, NA
        - \textbf{Relationship:} \\
            Family: [parent of, child of, spouse of, sibling of, fiancé of, distant family member of, grandparent of, grandchild of, uncle of, aunt of, nephew/niece of, cousin of, Other], \\
            Social: [neighbor of, friend of, lover of, ex-lover of, enemy of, idol of', 'member of same club as', 'Other',], \\
           \textbf{ Professional:} ['boss of', 'employee of', 'employer of','teacher of', 'student of','doctor of', 'patient of','seller to', 
                'client of','colleague of', 'classmate of','religious relationship with', 'Other'] \\    
        \textbf{Familiarity (A → B)}: [Familiar, Unfamiliar, NA] \\
        - \textbf{Intentions alignment (from the perspective of A):} [Aligned, Complementary,Conflicting, NA]   \\
        Aligned goals occur when both speakers share the same intentions, perspectives, or desired outcomes. Their interaction shows agreement, mutual understanding, or a common purpose. \\ 
        Complementary goals occur when the speakers’ intentions differ but support each other in a productive way. They may bring unique perspectives or roles—such as one generating ideas and the other evaluating them—while still working toward a shared or compatible outcome. \\
        Conflicting goals arise when the speakers’ aims or intentions oppose each other. This often happens in arguments, debates, or negotiations where one person’s goal contradicts the other’s.
        NA (Not Applicable) applies when the dialogue does not clearly reveal the speakers’ goals or when the interaction is neutral, purely informational, or casual. Example: A brief exchange where one speaker gives directions and the other simply acknowledges them. \\     
        -\textbf{ Intentions alignment (from your perspective):} [Aligned, Complementary,Conflicting, NA]   \\
        - \textbf{Overall power difference "powerDiff" (from the perspective of A):} ['High power', 'Equal power', 'Less power', 'Neutral']\\
        - \textbf{Overall power difference "powerDiffPersp" (from your perspective): }['High power', 'Equal power', 'Less power', 'Neutral'] \\
        - \textbf{Overall social status difference "socialDiff" (from the perspective of A): }['Higher status', 'Equal status', 'Lower status', 'Neutral'] \\ 
        - \textbf{Overall social status difference "socialDiffPersp" (from your perspective):}['Higher status', 'Equal status', 'Lower status', 'Neutral'] \\
   \textbf{2. You have to do the same, but from person B to A, and the remaining combinations.}\\     
\textbf{3. Intentions alignment}
For each pair of speakers, determine their intention alignment from your perspective. Intention alignment can be: Aligned, Complementary
        , Conflicting, NA. Answer NA if it is hard or not possible to determine. Example :  "Person A | Person B": "Conflicting" \\ 
\textbf{4. dialogue Context}\\
    1. Location domain: [Personal, professional ]\\
    2. Location privacy: [Private, Public ]\\ 
\textbf{**Overall summary**}:\{overall summary\}\\
**Scene description**: {scene details}\\
**Input Dialog**: {dialog}\\

\textbf{Instructions }\\
- Understand carefully what the movie is about, the current dialogue and scene. Based on your understanding, return:
- **Profiles**: a json list of profiles for the persons involved in the dialogue only. Make sure to only include the persons who are speaking in the dialogue and not the referenced \\
- **Speakers Dynamics**: a list of json containing speakers dynamics \\
- **Intention alignment**: a list of pairs of speakers and their alignment e.g: ["Omar, James":"Complementary", "Lee, Samy":"Conflicting"] \\ 
- **Dialogue Context**: a dict containing Social setting, Location domain, Location privacy, Dialogue formality. \\
- The output must be a JSON. 

\end{tcolorbox}
\end{figure}